\documentclass[letterpaper]{article} % DO NOT CHANGE THIS
\usepackage{aaai23}  % DO NOT CHANGE THIS
\usepackage{times}  % DO NOT CHANGE THIS
\usepackage{helvet}  % DO NOT CHANGE THIS
\usepackage{courier}  % DO NOT CHANGE THIS
\usepackage[hyphens]{url}  % DO NOT CHANGE THIS
\usepackage{graphicx} % DO NOT CHANGE THIS
\usepackage{natbib}  % DO NOT CHANGE THIS AND DO NOT ADD ANY OPTIONS TO IT
\usepackage{caption} % DO NOT CHANGE THIS AND DO NOT ADD ANY OPTIONS TO IT
\usepackage{algorithm}
\usepackage{algpseudocode}

\usepackage{newfloat}
\usepackage{listings}
\DeclareCaptionStyle{ruled}{labelfont=normalfont,labelsep=colon,strut=off} % DO NOT CHANGE THIS
\floatstyle{ruled}
\newfloat{listing}{tb}{lst}{}
\floatname{listing}{Listing}
\title{Latent Autoregressive Source Separation}
\author {
    Emilian Postolache\thanks{Equal contribution. Listing order is random.}\textsuperscript{\rm 1},
    Giorgio Mariani\textsuperscript{$*$\rm 1},
    Michele Mancusi\textsuperscript{$*$\rm 1},
    Andrea Santilli\textsuperscript{\rm 1},\\
    Luca Cosmo\thanks{Shared last authorship.}\textsuperscript{\rm 2},
    Emanuele Rodol\`a\textsuperscript{$\dagger$\rm 1}
}
\affiliations {
    \textsuperscript{\rm 1} Sapienza University of Rome, Italy\\
    \textsuperscript{\rm 2} Ca’ Foscari University of Venice, Italy\\
    postolache@di.uniroma1.it, mariani@di.uniroma1.it, mancusi@di.uniroma1.it
}

\usepackage{bibentry}
\usepackage{amsmath,amssymb}
\usepackage{multirow}
\usepackage{booktabs}
\usepackage{comment}
\usepackage{tikz}
\usepackage{pdfpages}

\def\LASS{\emph{LASS }}
\newcommand{\Concat}[2]{\operatorname{\texttt{concat}}(#1,#2)}
\newcommand{\Prior}[0]{\text{prior}}
\newcommand{\Likelihood}[0]{\text{likelihood}}
\newcommand{\Posterior}[0]{\text{posterior}}

\begin{document}

\maketitle

\begin{abstract}
Autoregressive models have achieved impressive results over a wide range of domains in terms of generation quality and downstream task performance. In the continuous domain, a key factor behind this success is the usage of quantized latent spaces (e.g., obtained via VQ-VAE autoencoders), which allow for dimensionality reduction and faster inference times. However, using existing pre-trained models to perform new non-trivial tasks is difficult since it requires additional fine-tuning or extensive training to elicit prompting. This paper introduces \LASS as a way to perform vector-quantized \textit{Latent Autoregressive Source Separation} (i.e., de-mixing an input signal into its constituent sources) without requiring additional gradient-based optimization or modifications of existing models. 
Our separation method relies on the Bayesian formulation in which the autoregressive models are the priors, and a discrete (non-parametric) likelihood function is constructed by performing frequency counts over latent sums of addend tokens. We test our method on images and audio with several sampling strategies (e.g., ancestral, beam search) showing competitive results with existing approaches in terms of separation quality while offering at the same time significant speedups in terms of inference time and scalability to higher dimensional data.
\end{abstract}

\section{Introduction}
\label{sec:introduction}
\begin{figure*}[ht]
    \centering
    %\begin{overpic}[width=0.8\textwidth]{img/lass}%
%    \end{overpic}
    \includegraphics[width=0.9\textwidth]{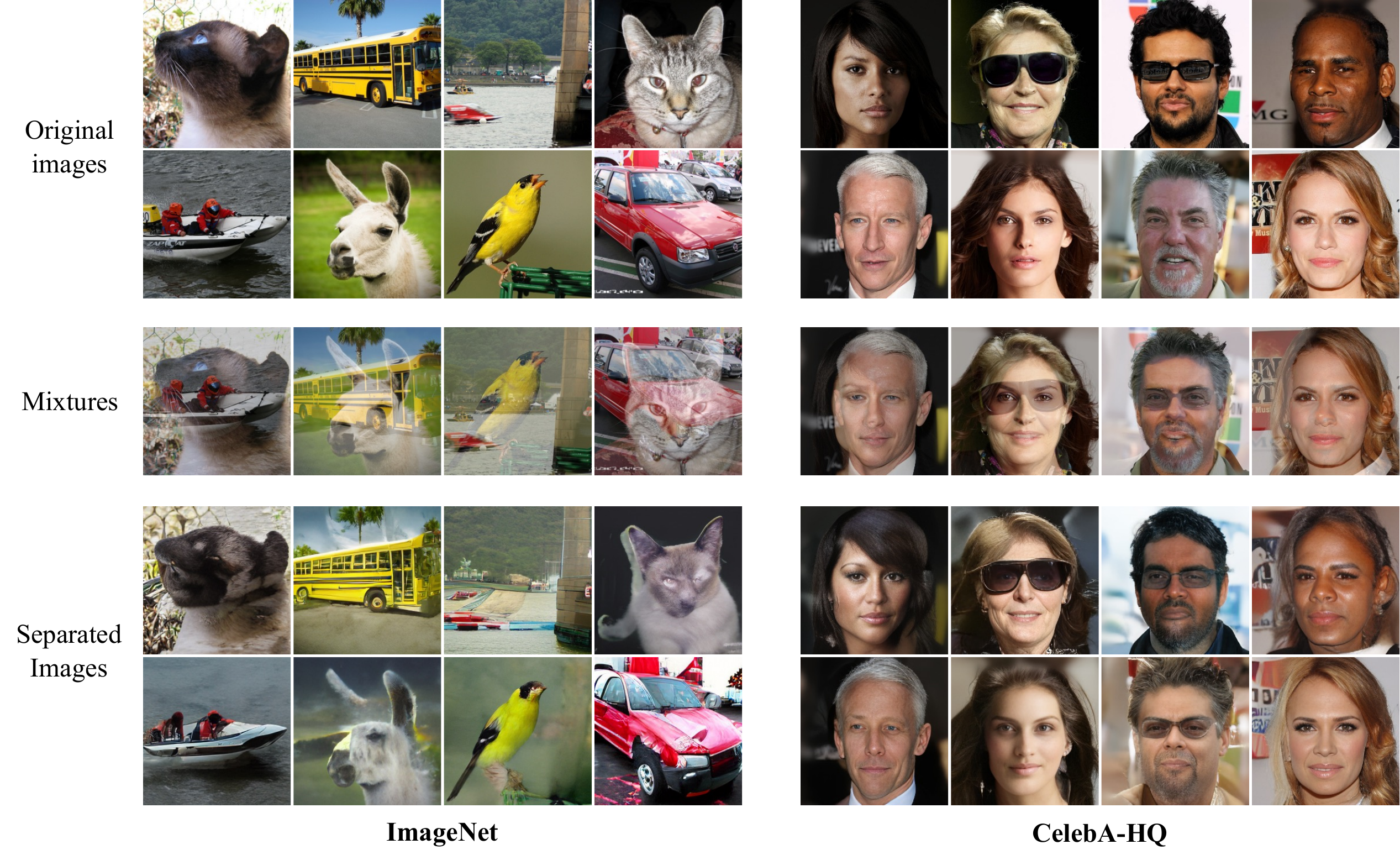}
    \caption{\label{fig:teaser}256x256 separations obtained with \LASS using pre-trained autoregressive models. Left: class-conditional ImageNet. Right: unconditional CelebA-HQ. }
\end{figure*}
Autoregressive models have achieved impressive results in a plethora of domains ranging from natural language \cite{brown2020language} to densely-valued domains such as audio \cite{dhariwal:2020} and vision \cite{razavi2019, esser2021taming}, including multimodal joint spaces \cite{ramesh2021zero, yu2022scaling}. 
In the dense setting, it is typical to train autoregressive models over discrete latent representations obtained through the quantization of continuous data, possibly using VQ-VAE autoencoders \cite{oord2017}. This way, generating higher resolution samples while simultaneously reducing inference time is possible.  Additionally, the learned latent representations are useful for downstream tasks \cite{castellon2021codified}. However, in order to perform new non-trivial tasks, the standard practice is to fine-tune the model or, in alternative, elicit prompting by scaling training  \cite{weiFLAN, sanh2022multitask}.
The former is usually the default option, but it requires additional optimization steps or modifications to the model. The latter is challenging on non-trivial tasks, especially in domains different from natural language \cite{https://doi.org/10.48550/arxiv.2208.02532, hertz2022prompt}.

This paper aims to tackle one of such tasks, namely \textit{source separation}, leveraging existing vector-quantized autoregressive models without requiring any gradient-based optimization or architectural modifications.
The task of separating two or more sources from a mixture signal has recently received much attention following the success of deep learning, especially in the audio domain, ranging from speech \cite{Dovrat2021ManySpeakersSC}, music \cite{defossez2021hybrid}, and universal source separation \cite{fuss, postolache2022adversarial}. Although not as prominent as its audio counterpart, image source separation has been addressed in literature \cite{Halperin2019}. Most successful approaches use explicit supervision to achieve notable results \cite{luo2019conv, defossez2019}, or leverage large-scale  unsupervised regression \cite{mixit}. 

We propose a generative approach to perform source separation via autoregressive prior distributions trained on a latent VQ-VAE domain (when class information is used, the approach is weakly supervised; otherwise, it is unsupervised). A non-parametric sparse likelihood function is learned by counting the occurrences of latent mixed tokens with respect to the sources' tokens, obtained by mapping the data-domain sum signals and the relative addends via the VQ-VAE. This module is not invasive, neither for the VQ-VAE nor for the autoregressive priors, given that the representation space of the VQ-VAE does not change while learning the likelihood function. Finally, the likelihood function is combined with the estimations of the autoregressive priors at inference time via the Bayes formula, resulting in a posterior distribution. The separations are obtained from the posterior distributions via standard discrete samplers (e.g., ancestral, beam search). We call our method \LASS \textit{(Latent Autoregressive Source Separation)}.

Our contributions are summarized as follows:
\begin{itemize}
    \item We introduce  \LASS as a Bayesian inference method for source separation  that can leverage existing pre-trained autoregressive models in quantized latent domains. 
    \item We experiment with \LASS in the image domain and showcase competitive results at a significantly smaller cost in inference time with respect to competitors on MNIST and CelebA (32$\times$32). We also showcase qualitative results on ImageNet (256$\times$256) and CelebA-HQ (256$\times$256), thanks to the scalability of \LASS to pre-trained models. To the best of our knowledge, this is the first method to scale generative source separation to higher resolution images.
    \item We experiment with \LASS in the music source separation task on the Slakh2100 dataset. \LASS obtains  performance comparable to state-of-the-art supervised models, with a significantly smaller cost in inference and training time with respect to generative competitors.
\end{itemize}

\section{Related Work}
\label{sec:related}
The problem of source separation has been classically tackled in an unsupervised fashion under the umbrella term of \textit{blind source separation} \cite{comon:1994, ica, rpca, Smaragdis:2014}. In this setting, there is no information regarding the sources to be separated from a mixture signal. As such, these methods rely on broad mathematical priors such as source independence \cite{ica} or repetition \cite{repet} to perform separation.
With the advent of deep learning, most prominent methods for source separation can be classified as regression-based or generative-based methods.

\subsection{Regression-based source separation}
In this setting, a mixture is fed to a parametric model (i.e., a neural network) that outputs the separated sources. Training is typically performed in a supervised manner by matching the estimated separations with the ground truth sources with a regression loss (e.g., $\mathcal{L}_1$ or $\mathcal{L}_2$) \cite{9746530}. Supervised regression has been applied to image source separation \cite{Halperin2019}, but it has been mainly investigated in the audio domain, where two approaches are prevalent: the mask-based approach and the waveform approach. In the mask-based approach, the model performs separation by applying estimated masks on mixtures, typically in the STFT domain \cite{Roweis:2000, Uhlich:2015, Huang2014SingingVoiceSF, 7492604, liu:2018,Takashi2018}.
In the waveform approach, the model outputs the estimated sources directly in the time domain to overcome phase estimation, which is required when transforming the signal from the STFT domain to the waveform domain \cite{DBLP:conf/interspeech/LluisPS19, luo2019conv, defossez2019}.

\subsection{Generative source separation}
Following the success of deep generative models \cite{goodfellow2014generative, Kingma2014, ho2020denoising, song2021scorebased}, a new class of generative source separation methods is gaining prominence. This setting emphasizes the exploitation of broad generative models (especially pre-trained ones) to solve the separation task without needing a specialized architecture (as with regression-based models). 

Following early work on deep generative separation based on GANs \cite{subakan2018generative, 10.5555/3367243.3367421, narayanaswamy2020unsupervised}, \citet{jayaram2020} propose the generative separation method BASIS in the image setting using score-based models \cite{song2019generative} (BASIS-NCSN) and a noise-annealed version of flow-based models (BASIS-Glow). The inference procedure is performed in the image domain through Langevin dynamics \cite{parisi}, obtaining good quantitative and qualitative results. The authors extend the Langevin dynamics inference procedure to autoregressive models by re-training them with a noise schedule, introducing the Parallel and Flexible (PnF) method \cite{jayaram2021parallel}. Although innovative, mainly when used for tasks such as inpainting, this method cannot use pre-trained autoregressive models directly, requiring fine-tuning under different noise levels. Further, working directly on the data domain, it exhibits a high inference time and scales with difficulty to higher resolutions. In this paper, we extend this line of research by proposing a separation procedure for latent autoregressive models that does not involve re-training, is scalable to arbitrary pre-trained checkpoints and is compatible with standard discrete samplers.

\section{Background}
\label{sec:background}

This section briefly introduces vector-quantized autoencoders (VQ-VAE) and autoregressive models, since they are core components of the separation procedure used in \textit{LASS}.

\subsection{VQ-VAE}

A data point $\mathbf{x} \in \mathbb{R}^N$ ($N$ is the total length of the data point, e.g., the length of the audio sequence or the number of pixel channels in an image) can be mapped to a discrete latent domain via a VQ-VAE \cite{oord2017}. First an encoder $E_\theta: \mathbb{R}^N \to \mathbb{R}^{S\times C}$ maps $\mathbf{x}$ to $E_\theta(\mathbf{x}) = (\mathbf{h}_{1}, \dots, \mathbf{h}_{S})$, where $C$ denotes the number of latent channels and $S$ the length of the latent sequence. A bottleneck block $B : \mathbb{R}^{S\times C} \to [K]^S$ casts the encoding into a discrete sequence $\mathbf{z} = (z_{1}, \dots, z_{S})$ by mapping each $\mathbf{h}_{s}$ into the index (also called token) $z_{s} = B(\mathbf{h}_{s})$  of the nearest neighbor $\mathbf{e}_{z_{s}}$ contained in an (ordered) set $\mathcal{C} = \{\mathbf{e}_k\}^K_{k=1}$ of learned vectors in $\mathbb{R}^C$ (called codes). A decoder $D_\psi:[K]^S \to \mathbb{R}^N$ maps the latent sequence back into the data domain, obtaining a reconstruction $\hat{\mathbf{x}} =  D_\psi(\mathbf{z})$.
VQ-GAN \cite{esser2021taming} is an enhanced version of the VQ-VAE, where the training loss is augmented with a discriminator and a perceptual loss, that improve reconstruction quality while increasing the compression rate of the autoencoder.
We refer the reader to \cite{oord2017} and \cite{esser2021taming} for more details on VQ-VAE and VQ-GAN. In the remainder of the article, we will refer to both models as VQ-VAE and make distinctions when necessary.

\subsection{Autoregressive models}

An autoregressive model defines a probability distribution over a discrete domain $[K]^S$ (in our case, the latent domain of the VQ-VAE). The joint probability of a sequence $\mathbf{z} = (z_1, \dots, z_S)$ is decomposed via the chain rule
\begin{align}
    p_\phi(\mathbf{z}) = \prod_{s = 1}^{S} p_\phi(z_{s}\vert \mathbf{z}_{< s}),
\end{align}
where $p_\phi(\cdot)$ is a learned parametric model, generally a neural network such as CNNs \cite{van2016conditional, salimans2017pixelcnn++} or Transformers \cite{vaswani2017attention}.
At inference time, samples can be obtained depending on the choice of a sampling procedure. Generally, ancestral sampling is used, where at each step, the token $z_{s}$ is drawn stochastically from the conditional $p_\phi(z_{s} \vert \mathbf{z}_{<s})$, possibly employing top-$k$ \cite{JMLR:v21:19-985} filtering to increase the diversity of the generated data \cite{Holtzman2020The}. When the goal is instead to maximize the probability of the whole sequence (w.r.t. all the sequences), heuristics like beam search are used \cite{reddy1977speechBeam}.
Beam search maintains $B$ possible hypotheses (beams) $\mathbf{z}^1, \dots, \mathbf{z}^B$ in parallel during inference. At each step $s$, it computes the conditional distributions $p_\phi(z^b_{s}\vert \mathbf{z}^b_{<s})$ for each beam and selects the $B$ new hypotheses that maximize the joint distributions $p_\phi(\mathbf{z}^b_{<s})p_\phi(z_{s}| \mathbf{z}^b_{<s})$.

\begin{figure*}[ht]
    \centering
    %\begin{overpic}[width=0.8\textwidth]{img/lass}%
%    \end{overpic}
    \includegraphics[width=0.85\textwidth]{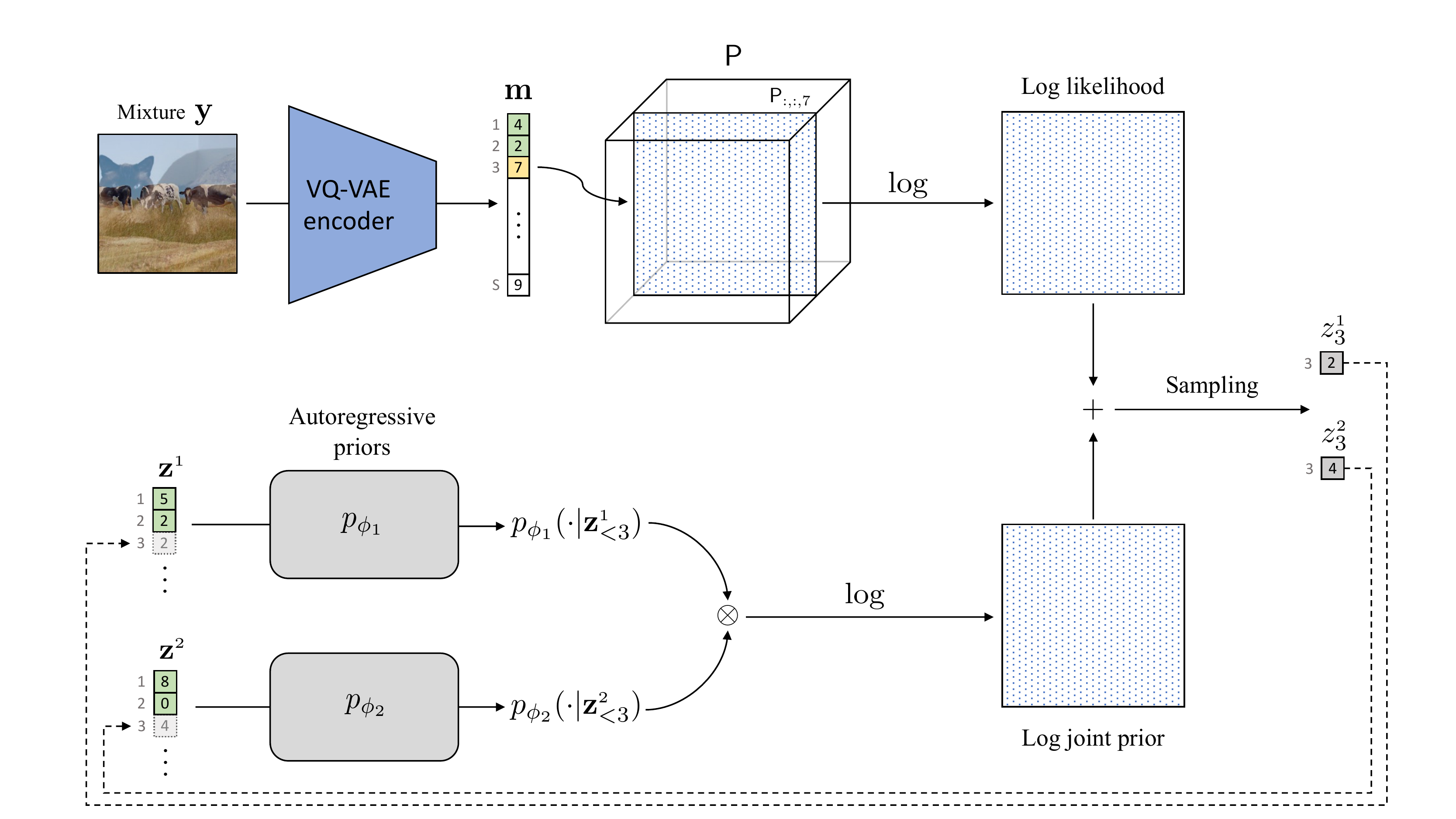}
    \caption{Schematic of the \LASS separation procedure. The picture shows the separation procedure at $s=3$ and is repeated until $s=S$. At the end of inference, we obtain $\mathbf{x}^{\scriptscriptstyle{1}}$ and $\mathbf{x}^{\scriptscriptstyle{2}}$ decoding $\mathbf{z}^{\scriptscriptstyle{1}}$ and $\mathbf{z}^{\scriptscriptstyle{2}}$ via the VQ-VAE decoder (not depicted in the picture). We refer the reader to Algorithm \ref{alg:lass} for more details.}
    \label{fig:lass}
\end{figure*}

\section{Method}
\label{sec:method}
Let $\mathbf{x} = (\mathbf{x}^{\scriptscriptstyle{1}},\mathbf{x}^{\scriptscriptstyle{2}}) \in \mathbb{R}^{2 \times N}$ denote two sources distributed according to $p_{\text{data}} = (p^{\scriptscriptstyle{1}}_{\text{data}}, p^{\scriptscriptstyle{2}}_{\text{data}})$ and $\mathbf{y} = (\mathbf{x}^{\scriptscriptstyle{1}} + \mathbf{x}^{\scriptscriptstyle{2}})/2$ an observable mixture. The goal of generative source separation is to estimate the sources $\mathbf{x}$ given the mixture $\mathbf{y}$, %and a model of $p_{\text{data}}$, 
using the Bayesian posterior (assuming independent sources):
\begin{align}
\label{eq:simple_bayes}
p(\mathbf{x}^{\scriptscriptstyle{1}}, \mathbf{x}^{\scriptscriptstyle{2}} \vert \mathbf{y}) \propto p^{\scriptscriptstyle{1}}_\text{data}(\mathbf{x}^{\scriptscriptstyle{1}})p^{\scriptscriptstyle{2}}_\text{data}(\mathbf{x}^{\scriptscriptstyle{2}})
p(\mathbf{y} \vert \mathbf{x}^{\scriptscriptstyle{1}}, \mathbf{x}^{\scriptscriptstyle{2}}). \end{align}
 Working directly with Eq. \eqref{eq:simple_bayes} in the continuous data domain is inefficient. To overcome this problem, we first model $p_{\text{data}}$ with autoregressive models in the latent space of a VQ-VAE. By changing the domain, we subsequentially redefine the likelihood function $p(\mathbf{y} \vert \mathbf{x}^{\scriptscriptstyle{1}}, \mathbf{x}^{\scriptscriptstyle{2}})$ such that no gradient-based optimization or model re-training is required.
 We address the first issue in the following subsection and the second in the subsequent one. We then describe how to perform inference using \LASS to separate data and propose a post-inference refinement procedure.

\subsection{Latent autoregressive source separation}
This paper explores the case in which $p_{\text{data}}$ is estimated by a unique autoregressive model $p_{\phi}$ for all the sources (unsupervised\footnote{Not to be confused with the unsupervised blind setting, i.e., in our unsupervised setting we have access to sources but we do not have class labels.}) and the case in which we have two independent ones, $p_{\phi} = (p_{\phi_1}, p_{\phi_2})$, for each of the two sources (weakly supervised), either in terms of class-conditioned or independently trained models. We will focus on this latter case in the following, since the former can be generalized setting $p_{\phi_1}=p_{\phi_2}$.

We denote the latent sources and mixtures, respectively, with $\mathbf{z} = (\mathbf{z}^{\scriptscriptstyle{1}},\mathbf{z}^{\scriptscriptstyle{2}}) = B(E_\theta(\mathbf{x}))$ and $\mathbf{m} = B(E_\theta(\mathbf{y}))$. The posterior distribution in Eq. \eqref{eq:simple_bayes} can be locally expressed in the latent domain as
\begin{align}
\label{eq:posterior}
p(\mathbf{z}_{s} \vert \mathbf{z}_{<s}, \mathbf{m}_{\leq s}) \propto p_{\phi}(\mathbf{z}_s \vert \mathbf{z}_{<s})  p(\mathbf{m}_{\leq s} \vert \mathbf{z}_{\leq s}),
\end{align}
for all $s = 1, \dots, S$. The first factor is the (joint) Bayesian prior, modeled with autoregressive distributions. The second factor is the likelihood function, which quantifies the likelihood of the sequences  $\mathbf{z}^{\scriptscriptstyle{1}}_{\leq s}, \mathbf{z}^{\scriptscriptstyle{2}}_{\leq s}$ to combine into $\mathbf{m}_{\leq s}$. 

Since each code in the convolutional VQ-VAE describes a local portion of the data, and given that the mixing operation is point-wise in the data domain,
the mixing relation between latent codes is local also in the latent domain. As such, we can drop the dependency on the previous context inside the likelihood function in Eq. \eqref{eq:posterior}, approximating it as
\begin{align}
\label{eq:likelihood_independency}
    p(\mathbf{m}_{\leq s} \vert \mathbf{z}_{\leq s}) 
\approx p(m_{s} \vert \mathbf{z}_s) .
\end{align}
Notice that not depending on the global context and thus on the specific position in the sequence, we can drop the position index $s$:
\begin{align}
\label{eq:index_drop}
    p(m_{s} \vert \mathbf{z}_s) = p(m_s \vert z^{{\scriptscriptstyle{1}}}_s, z^{{\scriptscriptstyle{2}}}_s) = p(m \vert z^{{\scriptscriptstyle{1}}}, z^{{\scriptscriptstyle{2}}}).
\end{align}

The following subsection describes how \LASS models the likelihood function.

\subsection{Discrete likelihood functions for source separation}
 Previous works in generative source separation \cite{jayaram2020, jayaram2021parallel} model likelihood functions directly in the data domain, typically employing a $\sigma$-isotropic Gaussian term 
 \begin{align}
     \label{eq:gaussian}
    p(\mathbf{y} \vert \mathbf{x}) = \mathcal{N}(\mathbf{y}
    \vert (\mathbf{x}^{{{\scriptscriptstyle{1}}}} + \mathbf{x}^{{{\scriptscriptstyle{2}}}})/2, \sigma^2 \mathbf{I}).
 \end{align}
 In our setting, we cannot combine $z^{{\scriptscriptstyle{1}}}_{s}$ and $z^{{\scriptscriptstyle{2}}}_{s}$ (or the associate dense codes $\mathbf{e}_{z^{{\scriptscriptstyle{1}}}_{s}}$ and $\mathbf{e}_{z^{{\scriptscriptstyle{2}}}_{s}}$) with the canonical sum operation, given that the VQ-VAE does not impose an explicit arithmetic structure on the latent space.
 
To cope with this, we model the likelihood function in Eq. \eqref{eq:index_drop} using discrete conditionals, represented with rank-$3$ tensors\footnote{We follow the notation for tensors as in \citet{Goodfellow-et-al-2016}.} $\mathsf{P} \in \mathbb{R}^{K \times K \times K}$:
\begin{align}
    p(\cdotp \vert z^{{\scriptscriptstyle{1}}}, z^{{\scriptscriptstyle{2}}})  = \mathsf{P}_{z^{{\scriptscriptstyle{1}}}, z^{{\scriptscriptstyle{2}}}, :}
\end{align}
In order to learn $\mathsf{P}$, we perform frequency counts on latent mixed tokens given the latent sources' tokens, by iterating over a dataset $X$. We first initialize a null integer tensor $\mathsf{F}^{0} \in \mathbb{N}^{K \times K \times K}$. Iterating over $\mathbf{x}^{{\scriptscriptstyle{1}}}, \mathbf{x}^{{\scriptscriptstyle{2}}} \in X$, we compute $\mathbf{y} = (\mathbf{x}^{{\scriptscriptstyle{1}}} + \mathbf{x}^{{\scriptscriptstyle{2}}})/2$, then obtain the latent sequences $\mathbf{z}^{{\scriptscriptstyle{1}}} = B( E_\theta(\mathbf{x}^{{\scriptscriptstyle{1}}})), \mathbf{z}^{{\scriptscriptstyle{2}}} = B( E_\theta(\mathbf{x}^{{\scriptscriptstyle{2}}}))$ and $\mathbf{m} = B( E_\theta(\mathbf{y}))$. For each entry $(z^{{\scriptscriptstyle{1}}}_s, z^{{\scriptscriptstyle{2}}}_s, m_s) \in  (\mathbf{z}^{{\scriptscriptstyle{1}}}, \mathbf{z}^{{\scriptscriptstyle{2}}}, \mathbf{m})$, at step $t$, we simply increment the previous count by one: \begin{align}
    \mathsf{F}^{t}_{z^{{\scriptscriptstyle{1}}}_{s},z^{{{\scriptscriptstyle{2}}}}_s,m_s} = \mathsf{F}^{t-1}_{z^{{\scriptscriptstyle{1}}}_{s},z^{{{\scriptscriptstyle{2}}}}_{s},m_{s}} + 1\,,
    \\
    \mathsf{F}^{t}_{z^{{\scriptscriptstyle{2}}}_{s},z^{{\scriptscriptstyle{1}}}_{s},m_s} = \mathsf{F}^{t-1}_{z^{{\scriptscriptstyle{2}}}_{s},z^{{\scriptscriptstyle{1}}}_{s},m_s} + 1\,.
\end{align}
We permute the order of the addends in order to enforce the commutative property of the sum. After performing the statistics, we can define $\mathsf{P}$ as:
%by iterating over $z^{{\scriptscriptstyle{1}}}, z^{{\scriptscriptstyle{2}}}$ normalizing the conditionals:
\begin{equation}
    \mathsf{P}_{z^{{\scriptscriptstyle{1}}}, z^{{\scriptscriptstyle{2}}}, :} = 
    \frac{1}{\sum_{m=1}^{K} \mathsf{F}_{z^{{\scriptscriptstyle{1}}}, z^{{\scriptscriptstyle{2}}}, m}} \mathsf{F}_{z^{{\scriptscriptstyle{1}}}, z^{{\scriptscriptstyle{2}}}, :}  \end{equation}
At inference time, the likelihood function (parametric in $z^{{\scriptscriptstyle{1}}}$ and $z^{{\scriptscriptstyle{2}}}$, with $m$ fixed) can be obtained by slicing the tensor along $m$, namely   
\begin{equation}
p(m \vert \cdot, \cdot )  = \mathsf{P}_{:, :, m}.
 \end{equation}

At first glance, modeling the conditional distributions without parameters could seem memory inefficient, with a complexity of $O(K^3)$. In practice, the tensor $\mathsf{P}$ is \textit{highly sparse}. We showcase this in Table \ref{tab:stat} for all our experiments, where the density of $\mathsf{P}$ is defined as the percentage of nonzero elements in $\mathsf{P}$.

Employing discrete likelihood functions for source separation in the latent domain of a VQ-VAE is a flexible approach; there is no need to change the VQ-VAE representation, the non-parametric learning procedure does not depend on hyperparameters, and the autoregressive priors do not require re-training.

\begin{algorithm}[t]
\caption{\LASS inference \label{alg:lass}}
\textbf{Input:} $\mathbf{y}$\\%, \tau$\\
\textbf{Output:} $\mathbf{x}^{{\scriptscriptstyle{1}}}, \mathbf{x}^{{\scriptscriptstyle{2}}}$
\begin{algorithmic}[1]
\State $\mathbf{m} \gets B(E_\theta(\mathbf{y}))$
%\State $\mathbf{z}_1, \mathbf{z}_2 \gets [\texttt{BOS}], [\texttt{BOS}]$
\State $\mathbf{z}^{{\scriptscriptstyle{1}}} \gets  \text{[]}$
\State $\mathbf{z}^{{\scriptscriptstyle{2}}} \gets \text{[]}$
\For{$s=1$ to $S$}
\State $\Prior{} \gets \log(p_{\phi_1}(\cdotp \vert \mathbf{z}^{{\scriptscriptstyle{1}}}) \otimes p_{\phi_2}(\cdotp \vert \mathbf{z}^{{\scriptscriptstyle{2}}}))$
\State $\Likelihood{} \gets \log( \mathsf{P}_{:, :, m_s})$
\State $\Posterior{}  \gets \Prior{} + \lambda\; \Likelihood{}$
%\State $z_{1,s}, z_{2,s} \gets \arg \min P$
\State $(z^{{\scriptscriptstyle{1}}}_s,z^{{\scriptscriptstyle{2}}}_s) \gets \texttt{Sampler}( \text{posterior})$ %\arg \min_{z^{{\scriptscriptstyle{1}}}, z^{{\scriptscriptstyle{2}}}} \Posterior{}[z^{{\scriptscriptstyle{1}}},z^{{\scriptscriptstyle{2}}}]$
\State $\mathbf{z}^{{\scriptscriptstyle{1}}} \gets \Concat{\mathbf{z}^{{\scriptscriptstyle{1}}}}{z^{{\scriptscriptstyle{1}}}_s}$
\State $\mathbf{z}^{{\scriptscriptstyle{2}}} \gets \Concat{\mathbf{z}^{{\scriptscriptstyle{2}}}}{z^{{\scriptscriptstyle{2}}}_s}$
\EndFor
\State $\mathbf{x}^{\scriptscriptstyle 1} \gets D_\psi(\mathbf{z}^{\scriptscriptstyle 1})$
\State $\mathbf{x}^{\scriptscriptstyle 2} \gets D_\psi(\mathbf{z}^{\scriptscriptstyle 2})$
\State \Return $\mathbf{x}^{\scriptscriptstyle 1}$, $\mathbf{x}^{\scriptscriptstyle 2}$
\end{algorithmic}
\end{algorithm}
 \subsection{Inference procedure} Given an observable mixture $\mathbf{y}$, the autoregressive priors $p_{\phi_1}, p_{\phi_2}$ and the learned likelihood tensor $\mathsf{P}$, it is possible to perform inference and estimate $\mathbf{x}^{{\scriptscriptstyle{1}}}, \mathbf{x}^{{\scriptscriptstyle{2}}}$, as described in Algorithm \eqref{alg:lass} and depicted in Figure \ref{fig:lass}. 
 
 We start by mapping $\mathbf{y}$ to the latent domain obtaining $\mathbf{m} = B(E_\theta(\mathbf{y}))$ and initializing the estimates $\mathbf{z}^{{\scriptscriptstyle{1}}}, \mathbf{z}^{{\scriptscriptstyle{2}}}$ with the empty sequences. The algorithm iterates over $s = 1, \dots, S$. 
 
 At each step, the joint prior (a $K \times K$ matrix) is computed (Line 5) by taking the outer product of the two distributions predicted by the autoregressive models conditioned over the past context. We use the logarithms of the distributions for numerical stability. The log-likelihood function is computed next (Line 6), applying the logarithm on $\mathsf{P}_{:,:,m_s}$.  In our experiments,  we can apply different scaling factors $\lambda$ to the log-likelihood to balance it to the priors. The two matrices are then combined to form the posterior on Line 7.
 
 Finally (Lines 8-10), different techniques can be employed to sample the best candidate tokens $(z^{{\scriptscriptstyle{1}}}_s, z^{{\scriptscriptstyle{2}}}_s)$ from the posterior. In our experiments, we used ancestral sampling (with and without top-$k$ filtering) and beam search. After the inference loop ends, the estimated sequences are mapped back to the data domain with the decoder of the VQ-VAE (Lines 12-13), obtaining $\mathbf{x}^{{\scriptscriptstyle{1}}}$ and $\mathbf{x}^{{\scriptscriptstyle{2}}}$.
 
 \subsubsection{Post-inference refinement}
The quality of the separated images is limited by the quality of the images generated by the VQ-VAE autoencoder. To enhance the separations we can adopt an additional refinement step by iteratively optimizing the VQ-VAE latent representations of the samples:
 \begin{align}
 \label{eq:refinement1}
 \mathbf{e}^{{\scriptscriptstyle{1}}}_{t+1} = \mathbf{e}^{{\scriptscriptstyle{1}}}_{t} + \alpha \nabla_{\mathbf{e}^{{\scriptscriptstyle{1}}}_t}  \Vert  D_\psi(\mathbf{e}^{{\scriptscriptstyle{1}}}_t) + D_\psi(\mathbf{e}^{{\scriptscriptstyle{2}}}_t) - 2\mathbf{y} \Vert_2 \\
  \label{eq:refinement2}
 \mathbf{e}^{{\scriptscriptstyle{2}}}_{t+1} =  \mathbf{e}^{{\scriptscriptstyle{2}}}_{t} + \alpha
 \nabla_{\mathbf{e}^{{\scriptscriptstyle{2}}}_t}
 \Vert  D_\psi(\mathbf{e}^{{\scriptscriptstyle{1}}}_t) + D_\psi(\mathbf{e}^{{\scriptscriptstyle{2}}}_t) - 2\mathbf{y} \Vert_2
\end{align} 
for $t= 1, \dots, T - 1$ and $\mathbf{e}^{{\scriptscriptstyle{1}}}_1 = E_\theta( \mathbf{x}^{{\scriptscriptstyle{1}}})$, $\mathbf{e}^{{\scriptscriptstyle{2}}}_1 = E_\theta(\mathbf{x}^{{\scriptscriptstyle{2}}})$. In simple words, we optimize for dense latent embeddings such that their decodings better sum to the mixture, initializing them to the output of Algorithm \ref{alg:lass}.
We found this strategy particularly helpful on the MNIST datset, where we assess the quality of the separation through a pixel-wise metric (PSNR) and the VQ-VAE tends to produce smooth images. 

\section{Experiments}
\label{sec:experiments}
We perform quantitative and qualitative experiments on various datasets to demonstrate the efficacy and scalability of \emph{LASS}.
In the image domain, we evaluate on MNIST \cite{mnist} and CelebA (32$\times$32) \cite{liu2015faceattributes} and present qualitative results on the higher resolution datasets CelebA-HQ (256$\times$256) \cite{karras2018progressive} and ImageNet (256$\times$256) \cite{imagenet}. In the audio domain, we test on Slakh2100 \cite{manilow2019}, a large dataset for music source separation suitable for generative modeling. We conducted all our experiments on a single Nvidia RTX 3090 GPU with 24 GB of VRAM. Implementation details for all the models are listed on the companion website\footnote{\texttt{github.com/gladia-research-group/\\latent-autoregressive-source-separation}}.

\subsection{Image source separation}

\begin{table}[t]
  \centering
  \begin{tabular}{llll}
    \toprule
        Dataset &  $K$     &  Density (\%)  \\
    \midrule
    MNIST & 256 &  $1.49\times10^0$ \\
    CelebA & 512 &  $6.06\times10^0$ \\
    CelebA-HQ & 1024 & $3.80\times10^{-1}$ \\
    ImageNet & 16384 &  $3.90\times10^{-3}$ \\
    %\midrule 
    SLAKH (Drum + Bass) & 2048  & $7.60\times10^{-2}$ \\
    \bottomrule
  \end{tabular}
  \caption{  \label{tab:stat}Statistics on likelihood functions over different datasets. $K$ is the number of VQ-VAE (or VQ-GAN) latent codes. Density is the percentage of nonzero elements in the likelihood function.}
\end{table}

We choose the Transformer architecture \cite{vaswani2017attention} as the autoregressive backbone for all image source separation experiments. With MNIST and CelebA, we first train a VQ-VAE, then train the autoregressive Transformer on its latent space. We use $K=256$ codes on MNIST and $K=512$ on CelebA, given that CelebA presents more variability, requiring more information to reconstruct data. On CelebA-HQ and ImageNet, we leverage pre-trained VQ-GANs \cite{esser2021taming} alongside the pre-trained Transformers published by the authors\footnote{ \texttt{github.com/CompVis/taming-transformers}} (\texttt{celebahq\_transformer} checkpoint for CelebA-HQ and \texttt{cin\_transformer} for ImageNet).
Given the flexibility of \LASS, they are employed inside the separation algorithm without modifications. On CelebA-HQ the VQ-GAN has $K=1024$ codes, while on ImageNet has $K=16384$. As a first step, in all image-based experiments we learn the $\mathsf{P}$ tensor using the procedure presented in the section ``Method". As shown in Table \ref{tab:stat}, CelebA presents the lowest sparsity (highest density) while ImageNet has the highest. In all cases, density is below $7\%$, and the inference procedure is not affected by memory issues.

\begin{figure}[t]
    \centering
    %\begin{overpic}[width=0.8\textwidth]{img/lass}%
%    \end{overpic}
    \includegraphics[width=1\columnwidth]{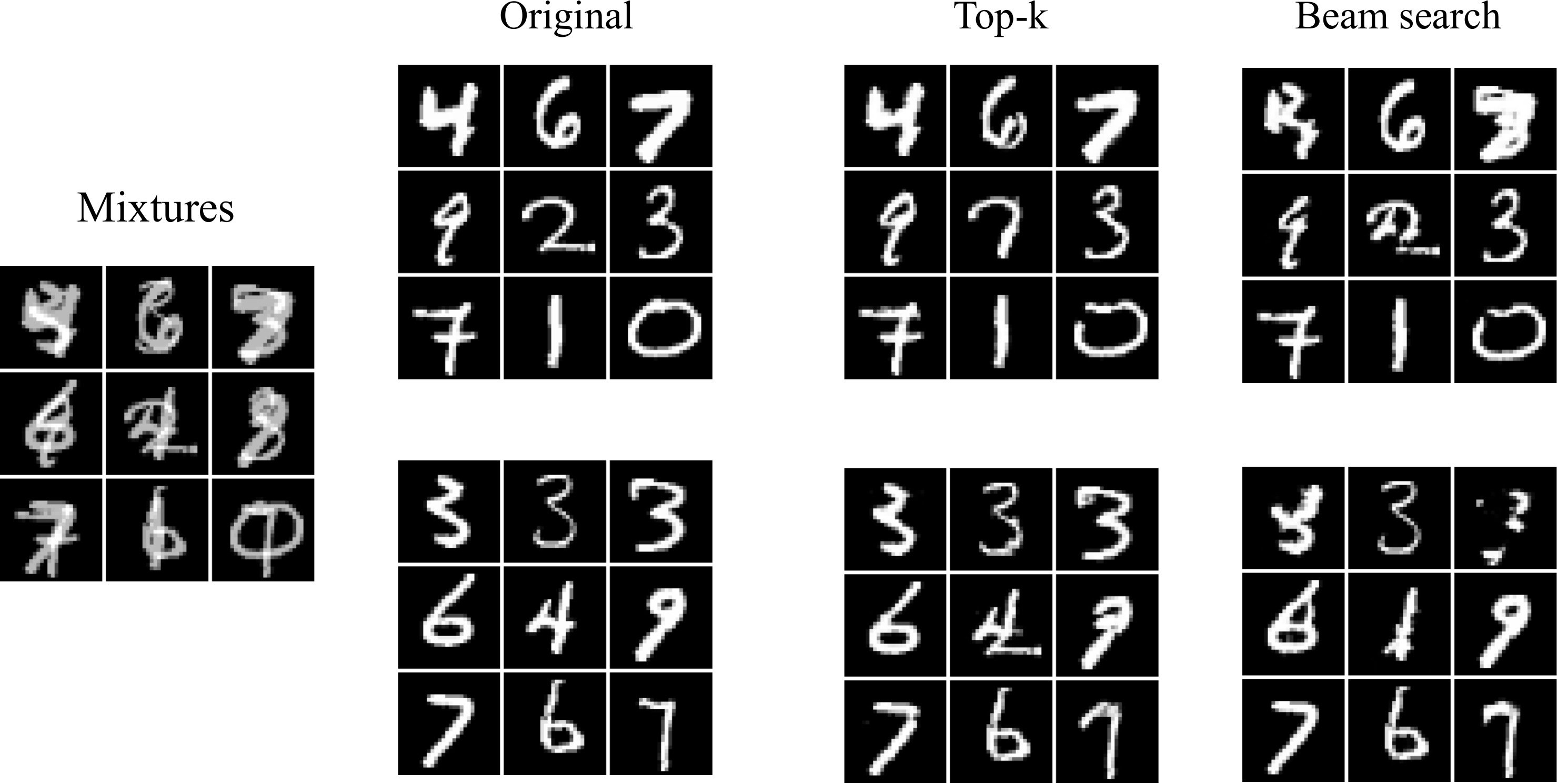}
    \caption{\label{fig:mnist}Results on MNIST with top-$k$ sampling ($k = 32$) over a random batch of examples. Top-$k$ sampling produces more defined digits, in agreement with the results in Table \ref{tab:sampler}.}
\end{figure}
\subsubsection{Quantitative results}
\begin{table}
\centering
\begin{tabular}{lll}\toprule
Separation Method & MNIST (PSNR)  & CelebA (FID)\\\midrule
Average           & 14.9 &  15.19 \\
NMF               & 9.4  & -\\
S-D               & 18.5 & -\\
BASIS Glow      & 22.7 & -\\
BASIS NCSN      & {29.3} & 7.55\\
\textbf{\LASS (Ours)}     & 24.2 & {8.96}  \\\bottomrule
\end{tabular}
\caption{\label{tab:images}Comparison with other methods on MNIST and CelebA test set. Results are reported in PSNR (higher is better) and FID (lower is better). } 
\end{table}

To assess the quality of image separations produced by \emph{LASS}, we compare our method with different baselines on MNIST and CelebA.

On MNIST, we compare \LASS with results reported for the two generative separation methods  ``BASIS NCSN" (score-based) and  ``BASIS Glow" (noise-annealed flow-based) from \cite{jayaram2020}, the GAN-based ``S-D" method \cite{10.5555/3367243.3367421}, the fully supervised version of Neural Egg  ``NES"  and the “Average” baseline, where separations are obtained directly from the mixture $\mathbf{x}^{{\scriptscriptstyle{1}}} = \mathbf{x}^{{\scriptscriptstyle{2}}} = \mathbf{y} / 2$. In all these cases, the evaluation metric is the PSNR (Peak Signal to Noise Ration) \cite{5596999}. We follow the experimental procedure of \cite{jayaram2020} on MNIST and perform separation on a set of 6,000 mixtures obtained by combining 12,000 test sources.
In order to choose the best sampler for this dataset, we validate the set of samplers in Table \ref{tab:sampler} on 1,000 mixtures constructed from the test split. We find that stochastic samplers perform best (PSNR $>$ 20 dB) while MAP methods do not reach a satisfactory performance. We hypothesize that beam search tends to fall into sub-optimal solutions by performing incorrect choices in early inference over sparse images such as MNIST digits. Top-$k$ sampling with $k=32$ performs best, so we choose it to perform the evaluation (a qualitative comparison is shown in Figure \ref{fig:mnist}). For each mixture in the test set we sample a candidate batch of 512 separations, select the separation whose sum better matches the mixture (w.r.t. the $\mathcal{L}_2$ distance), and finally perform the refinement procedure in Eqs. \eqref{eq:refinement1}, \eqref{eq:refinement2} with $T=500$ and $\alpha=0.1$.
Evaluation metrics on this experiment are shown in Table \ref{tab:images}, while inference time is reported in Table \ref{tab:speed}. Our method achieves higher metrics than ``NMF", ``S-D" and ``BASIS Glow" and is faster than ``BASIS NCSN", thanks to the latent quantization. The higher PSNR achieved by the later method can be attributed to the fact that, in their case, the underlying generative models perform sampling directly in the image domain; in our case, the VQ-VAE compression can hinder the metrics.

We compare our method to ``BASIS NCSN", using the pre-trained NCSN model \cite{song2019generative} on CelebA. In this case, we evaluate against the FID metric \cite{NIPS2017_8a1d6947} instead of PSNR, given that for datasets that feature more variability than MNIST, source separation can be an underdetermined task \cite{jayaram2020}: semantically good separations can receive a low PSNR score since the generative models may alter features such as color and cues (an effect amplified by a VQ-GAN decoder). The FID metric better quantifies if the separations belong to the distribution of the sources. We test on 10,000 mixtures computed from pair of images in the validation split using a top-$k$ sampler with $k = 32$. We scale the likelihood term by multiplying it by $\lambda = 3$. It is a known fact in the literature that score-based models outperform autoregressive models on FID metrics \cite{Dockhorn2021ScoreBasedGM} on different datasets, yet our method paired with an autoregressive model shows competitive results with respect to the score-based ``BASIS NCSN". 
\begin{table}[t]
  \centering
  \begin{tabular}{lll}
    %\toprule
    %&  \multicolumn{4}{c}{Reconstruction metrics}         \\
    %\cmidrule(r){2-6}

    \toprule
    Sampling Method   & MNIST (PSNR)  & SLAKH (SDR)  \\
    \midrule
     Greedy  & 17.36 $\pm{}$ 5.90    & 1.23 $\pm{}$ 2.33  \\
      Beam Search  & 16.96 $\pm{}$ 5.78    &  5.01 $\pm{}$ 2.39\\
    %Diverse Beam Search   & 16.95 $\pm{}$ 5.77   & - \\
    Ancestral Sampl.   & 24.03 $\pm{}$ 6.37   & 4.23 $\pm{}$ 2.29 \\
    Top-$k$ ($k = 16$)   & 23.74 $\pm{}$ 6.55    & 3.13 $\pm{}$ 2.53\\
    Top-$k$ ($k = 32$)   & 24.23 $\pm{}$  6.23    & 2.93 $\pm{}$ 2.20 \\
    Top-$k$ ($k = 64$)   & 23.85 $\pm{}$  6.13    & 3.24 $\pm{}$ 3.29  \\ 
      \bottomrule

  \end{tabular}
  \caption{\label{tab:sampler}Performance of \LASS with different sampling methods. On MNIST, the reported score is PSNR (dB) (higher is better), while on SLAKH is SDR (dB) (higher is better). When stochastic samplers are used (ancestral or top-$k$), the selected solution in the batch is the one whose sum minimizes the $\mathcal{L}_2$ distance to the input mixture.}
\end{table}

\subsubsection{Qualitative results}
To demonstrate the flexibility of \LASS in using existing models without any modification, we leverage pre-trained checkpoints on CelebA-HQ and ImageNet. In this case, only the likelihood tensor $\mathsf{P}$ is learned.
We showcase a curated results list in Figure \ref{fig:teaser} and a more extensive list on the companion website. To the best of our knowledge, our method is the first to scale up to 256$\times$256 resolutions and can be used with more powerful latent autoregressive models without re-training (which is cumbersome for very large models). As such, end-users can perform generative separation without having access to extensive computational resources for training these large models. 

\begin{table}[t]
  \centering
  \begin{tabular}{llll}
    \toprule
         & Method      & Time   \\
    \midrule
    \multirow{2}{*}{MNIST}     & \textbf{\LASS (Ours)}     & 4.49 s $\pm$ 0.27 s     \\
        & BASIS NCSN    &  53.34 s $\pm$ 0.51 s   \\
    \midrule
    \multirow{2}{*}{SLAKH} & \textbf{\LASS (Ours)} & 1.33 min $\pm$  0.87 s \\
         & PnF &  42.29 min $\pm$ 1.08 s  \\
    \bottomrule
  \end{tabular}
  \caption{\label{tab:speed}Inference speed comparisons for computing one separation. To estimate variance, we repeat inference 10 times on MINST and 3 times on SLAKH. We consider 3-second-long mixtures on SLAKH.}
\end{table}

\begin{table}[t]
\centering
\begin{tabular}{lllll}\toprule
Separation Method & Avg & Drums & Bass  \\\midrule
rPCA  & { }0.82   & { }0.60 & { }1.05    \\
ICA   & -1.26 & -0.99 & -1.53    \\
HPSS   & -0.45 & -0.56 & -0.33     \\
REPET  & { }1.04  & { }0.53 & { }1.54    \\
FT2D  & { }0.95  & { }0.59 & { }1.31    \\ \midrule 
\textbf{\LASS (Ours)}   & { }4.86 & { }4.73 & { }4.98\\ \midrule
 Demucs & { }5.39 & { }5.42 & { }5.36  \\
Conv-Tasnet & { }5.47 & { }5.51 & { }5.43 \\\bottomrule
\end{tabular}
\caption{Comparison with other source separation methods on SLAKH (``Drums'' and ``Bass'' classes). Results are reported in SDR (dB) (higher is better). Lower part of the table shows supervised methods. With ``Avg" we refer to the mean between the results over the two classes.} 
\label{tab:SLAKH}
\end{table}

\subsection{Music source separation}
We perform experiments on the SLAKH2100 dataset \cite{manilow2019} for the music source separation task. This dataset contains 2100 songs with separated sources belonging to 34 instrument categories, for a total of 145 hours of mixtures. We focus on the ``Drums'' and ``Bass'' data classes, with tracks sampled at 22kHz. We use the public checkpoint of \citet{dhariwal:2020} for the VQ-VAE model, taking advantage of its expressivity in modeling audio data over a quantized domain. Given that such a model is trained at 44kHz, we upsample input data linearly, then downsample the output back at 22kHz. For the two autoregressive priors, we train two Transformer models, one for ``Drums'' and another for ``Bass'' and learn the likelihood function over the VQ-VAE (statistics are reported in Table \ref{tab:stat}). We compare \LASS to a set of unsupervised blind source separation methods -``rPCA" \cite{rpca}, ``ICA" \cite{ica}, ``HPSS" \cite{repet}, ``FT2D" \cite{ft2d} - and to two supervised baselines Demucs \cite{defossez2019} and Conv-Tasnet \cite{luo2019conv} using the SDR (dB) evaluation metric computed with the \texttt{museval} library \cite{SiSEC18}. To evaluate the methods, we select 900 music chunks of 3 seconds from the test splits of the ``Drums'' and ``Bass'' classes, combining them to form 450 mixtures. The validation dataset is constructed similarly (with different music chunks).
As a sampling strategy, we use beam search since it shows the best results on a validation of 50 mixtures (Table \ref{tab:sampler}), using $B=100$ beams. Evaluation results are reported in Table \ref{tab:SLAKH}: \LASS clearly performs better than all the blind unsupervised baselines and is comparable with the results obtained by methods that use supervision. Furthermore, we compare the time performance of \LASS against the generative source separation method ``PnF" \cite{jayaram2021parallel} by evaluating the time required to separate a mixture of 3 seconds sampled at 22 kHz (piano vs. voice on ``PnF"). Results in Table \ref{tab:speed} show that \LASS is significantly faster, and as such, it can be adopted in more realistic inference scenarios.

\section{Limitations}
%The proposed method has some limitations that should be addressed with future research. 
In this paper we limit our analysis to the separation of two sources. Even if this is a common setup especially in image separation \cite{jayaram2021parallel, Halperin2019}, dealing with multiple sources is a possible line of future work. %First, the method is limited to two source separation since.
Under our framework, this would require to increase the dimensions of the discrete distributions (both the priors and the likelihood function). To alleviate this problem, techniques such as recursive separation may be employed \cite{takahashi2019recursive}. 

Another limitation of the proposed method is the locality assumption taken in Eq. \eqref{eq:likelihood_independency}. Different tasks such as colorization and super-resolution would require a larger conditioning context, and newer quantization schemes to aggregate latent codes on global contexts (using self-attention in the encoder and the decoder of the VQ-VAE) \cite{yu2021vector}. Adopting a VQ-VAE quantized with respect to the latent channels \cite{xu2021anytime} combined with a parametric likelihood function could be a way to solve this limitation, while still maintaining the flexible separation between VQ-VAE, priors, and likelihoods presented in the paper.

\section{Conclusion}In this paper, we proposed \LASS as a source separation method for latent autoregressive models that does not modify the structure of the priors. We have tested our method on different datasets and have shown results comparable to state-of-the-art methods while being more scalable and faster at inference time. Additionally, we have shown qualitative results at a higher resolution than those proposed by the competitors. We believe our method will benefit from the improved quality of newer autoregressive models, improving both the quantitative metrics and the perceptive results. 

\pagebreak
\section*{Acknowledgments}
We thank Marco Fumero for helping to compute the blind unsupervised baseline metrics in the audio setting. This work is supported by the ERC Grant no. 802554 (SPECGEO) and the IRIDE grant from DAIS, Ca' Foscari university of Venice, Italy.

\bibliography{bibliography}

\begin{thebibliography}{64}
\providecommand{\natexlab}[1]{#1}

\bibitem[{Brown et~al.(2020)Brown, Mann, Ryder, Subbiah, Kaplan, Dhariwal,
  Neelakantan, Shyam, Sastry, Askell et~al.}]{brown2020language}
Brown, T.; Mann, B.; Ryder, N.; Subbiah, M.; Kaplan, J.~D.; Dhariwal, P.;
  Neelakantan, A.; Shyam, P.; Sastry, G.; Askell, A.; et~al. 2020.
\newblock Language models are few-shot learners.
\newblock \emph{Proc. NeurIPS}, 33: 1877--1901.

\bibitem[{Castellon, Donahue, and Liang(2021)}]{castellon2021codified}
Castellon, R.; Donahue, C.; and Liang, P. 2021.
\newblock Codified audio language modeling learns useful representations for
  music information retrieval.
\newblock \emph{arXiv preprint arXiv:2107.05677}.

\bibitem[{Comon(1994)}]{comon:1994}
Comon, P. 1994.
\newblock {Independent Component Analysis, a new concept?}
\newblock \emph{{Signal Processing}}.

\bibitem[{D{\'e}fossez(2021)}]{defossez2021hybrid}
D{\'e}fossez, A. 2021.
\newblock Hybrid Spectrogram and Waveform Source Separation.
\newblock In \emph{Proceedings of the ISMIR 2021 Workshop on Music Source
  Separation}.

\bibitem[{Deng et~al.(2009)Deng, Dong, Socher, Li, Li, and Fei-Fei}]{imagenet}
Deng, J.; Dong, W.; Socher, R.; Li, L.-J.; Li, K.; and Fei-Fei, L. 2009.
\newblock ImageNet: A large-scale hierarchical image database.
\newblock In \emph{Proc. CVPR}, 248--255.

\bibitem[{Dhariwal et~al.(2020)Dhariwal, Jun, Payne, Kim, Radford, and
  Sutskever}]{dhariwal:2020}
Dhariwal, P.; Jun, H.; Payne, C.; Kim, J.~W.; Radford, A.; and Sutskever, I.
  2020.
\newblock Jukebox: A Generative Model for Music.
\newblock arXiv:2005.00341.

\bibitem[{Dockhorn, Vahdat, and Kreis(2021)}]{Dockhorn2021ScoreBasedGM}
Dockhorn, T.; Vahdat, A.; and Kreis, K. 2021.
\newblock Score-Based Generative Modeling with Critically-Damped Langevin
  Diffusion.
\newblock \emph{ArXiv}, abs/2112.07068.

\bibitem[{Dovrat, Nachmani, and Wolf(2021)}]{Dovrat2021ManySpeakersSC}
Dovrat, S.; Nachmani, E.; and Wolf, L. 2021.
\newblock Many-Speakers Single Channel Speech Separation with Optimal
  Permutation Training.
\newblock In \emph{Interspeech}.

\bibitem[{Défossez et~al.(2019)Défossez, Usunier, Bottou, and
  Bach}]{defossez2019}
Défossez, A.; Usunier, N.; Bottou, L.; and Bach, F. 2019.
\newblock Music {Source} {Separation} in the {Waveform} {Domain}.
\newblock \emph{arXiv:1911.13254 [cs, eess, stat]}.
\newblock ArXiv: 1911.13254.

\bibitem[{Esser, Rombach, and Ommer(2021)}]{esser2021taming}
Esser, P.; Rombach, R.; and Ommer, B. 2021.
\newblock Taming transformers for high-resolution image synthesis.
\newblock In \emph{Proc. CVPR}, 12873--12883.

\bibitem[{Goodfellow, Bengio, and Courville(2016)}]{Goodfellow-et-al-2016}
Goodfellow, I.; Bengio, Y.; and Courville, A. 2016.
\newblock \emph{Deep Learning}.
\newblock MIT Press.
\newblock \url{http://www.deeplearningbook.org}.

\bibitem[{Goodfellow et~al.(2014)Goodfellow, Pouget-Abadie, Mirza, Xu,
  Warde-Farley, Ozair, Courville, and Bengio}]{goodfellow2014generative}
Goodfellow, I.; Pouget-Abadie, J.; Mirza, M.; Xu, B.; Warde-Farley, D.; Ozair,
  S.; Courville, A.; and Bengio, Y. 2014.
\newblock Generative adversarial nets.
\newblock \emph{Proc. NIPS}, 27.

\bibitem[{Gusó et~al.(2022)Gusó, Pons, Pascual, and Serrà}]{9746530}
Gusó, E.; Pons, J.; Pascual, S.; and Serrà, J. 2022.
\newblock On Loss Functions and Evaluation Metrics for Music Source Separation.
\newblock In \emph{Proc. ICASSP}, 306--310.

\bibitem[{Halperin, Ephrat, and Hoshen(2019)}]{Halperin2019}
Halperin, T.; Ephrat, A.; and Hoshen, Y. 2019.
\newblock Neural separation of observed and unobserved distributions.
\newblock \emph{36th International Conference on Machine Learning, ICML 2019},
  2019-June: 4548--4557.

\bibitem[{Hertz et~al.(2022)Hertz, Mokady, Tenenbaum, Aberman, Pritch, and
  Cohen-Or}]{hertz2022prompt}
Hertz, A.; Mokady, R.; Tenenbaum, J.; Aberman, K.; Pritch, Y.; and Cohen-Or, D.
  2022.
\newblock Prompt-to-Prompt Image Editing with Cross Attention Control.
\newblock \emph{arXiv preprint arXiv:2208.01626}.

\bibitem[{Heusel et~al.(2017)Heusel, Ramsauer, Unterthiner, Nessler, and
  Hochreiter}]{NIPS2017_8a1d6947}
Heusel, M.; Ramsauer, H.; Unterthiner, T.; Nessler, B.; and Hochreiter, S.
  2017.
\newblock GANs Trained by a Two Time-Scale Update Rule Converge to a Local Nash
  Equilibrium.
\newblock In \emph{Proc. NeurIPS}, volume~30.

\bibitem[{Ho, Jain, and Abbeel(2020)}]{ho2020denoising}
Ho, J.; Jain, A.; and Abbeel, P. 2020.
\newblock Denoising diffusion probabilistic models.
\newblock \emph{Proc. NeurIPS}, 33: 6840--6851.

\bibitem[{Holtzman et~al.(2020)Holtzman, Buys, Du, Forbes, and
  Choi}]{Holtzman2020The}
Holtzman, A.; Buys, J.; Du, L.; Forbes, M.; and Choi, Y. 2020.
\newblock The Curious Case of Neural Text Degeneration.
\newblock In \emph{Proc. ICLR}.

\bibitem[{Horé and Ziou(2010)}]{5596999}
Horé, A.; and Ziou, D. 2010.
\newblock Image Quality Metrics: PSNR vs. SSIM.
\newblock In \emph{Proc. ICPR}, 2366--2369.

\bibitem[{Huang et~al.(2012)Huang, Chen, Smaragdis, and
  Hasegawa-Johnson}]{rpca}
Huang, P.-S.; Chen, S.~D.; Smaragdis, P.; and Hasegawa-Johnson, M. 2012.
\newblock Singing-voice separation from monaural recordings using robust
  principal component analysis.
\newblock In \emph{Proc. ICASSP}, 57--60. IEEE.

\bibitem[{Huang et~al.(2014)Huang, Kim, Hasegawa-Johnson, and
  Smaragdis}]{Huang2014SingingVoiceSF}
Huang, P.-S.; Kim, M.; Hasegawa-Johnson, M.~A.; and Smaragdis, P. 2014.
\newblock Singing-Voice Separation from Monaural Recordings using Deep
  Recurrent Neural Networks.
\newblock In \emph{Proc. ISMIR}.

\bibitem[{Hyv{\"a}rinen and Oja(2000)}]{ica}
Hyv{\"a}rinen, A.; and Oja, E. 2000.
\newblock Independent component analysis: algorithms and applications.
\newblock \emph{Neural networks}, 13(4-5): 411--430.

\bibitem[{Jayaram and Thickstun(2020)}]{jayaram2020}
Jayaram, V.; and Thickstun, J. 2020.
\newblock Source Separation with Deep Generative Priors.
\newblock In \emph{Proc. ICML}, PMLR.

\bibitem[{Jayaram and Thickstun(2021)}]{jayaram2021parallel}
Jayaram, V.; and Thickstun, J. 2021.
\newblock Parallel and flexible sampling from autoregressive models via
  langevin dynamics.
\newblock In \emph{Proc. ICML}, 4807--4818. PMLR.

\bibitem[{Karras et~al.(2018)Karras, Aila, Laine, and
  Lehtinen}]{karras2018progressive}
Karras, T.; Aila, T.; Laine, S.; and Lehtinen, J. 2018.
\newblock Progressive Growing of {GAN}s for Improved Quality, Stability, and
  Variation.
\newblock In \emph{Proc. ICLR}.

\bibitem[{Kingma and Welling(2014)}]{Kingma2014}
Kingma, D.~P.; and Welling, M. 2014.
\newblock Auto-Encoding Variational Bayes.
\newblock In \emph{Proc. ICLR}.

\bibitem[{Kong et~al.(2019)Kong, Xu, Wang, Jackson, and
  Plumbley}]{10.5555/3367243.3367421}
Kong, Q.; Xu, Y.; Wang, W.; Jackson, P. J.~B.; and Plumbley, M.~D. 2019.
\newblock Single-Channel Signal Separation and Deconvolution with Generative
  Adversarial Networks.
\newblock In \emph{Proc. IJCAI}, 2747–2753. AAAI Press.
\newblock ISBN 9780999241141.

\bibitem[{Kool, van Hoof, and Welling(2020)}]{JMLR:v21:19-985}
Kool, W.; van Hoof, H.; and Welling, M. 2020.
\newblock Ancestral Gumbel-Top-k Sampling for Sampling Without Replacement.
\newblock \emph{Journal of Machine Learning Research}, 21(47): 1--36.

\bibitem[{Lecun et~al.(1998)Lecun, Bottou, Bengio, and Haffner}]{mnist}
Lecun, Y.; Bottou, L.; Bengio, Y.; and Haffner, P. 1998.
\newblock Gradient-based learning applied to document recognition.
\newblock \emph{Proceedings of the IEEE}, 86(11): 2278--2324.

\bibitem[{Liu and Yang(2018)}]{liu:2018}
Liu, J.-Y.; and Yang, Y.-H. 2018.
\newblock Denoising Auto-encoder with Recurrent Skip Connections and Residual
  Regression for Music Source Separation.
\newblock arXiv:1807.01898.

\bibitem[{Liu et~al.(2015)Liu, Luo, Wang, and Tang}]{liu2015faceattributes}
Liu, Z.; Luo, P.; Wang, X.; and Tang, X. 2015.
\newblock Deep Learning Face Attributes in the Wild.
\newblock In \emph{Proc. ICCV}.

\bibitem[{Lluís, Pons, and Serra(2019)}]{DBLP:conf/interspeech/LluisPS19}
Lluís, F.; Pons, J.; and Serra, X. 2019.
\newblock End-to-End Music Source Separation: Is it Possible in the Waveform
  Domain?
\newblock In \emph{INTERSPEECH}, 4619--4623.

\bibitem[{Luo and Mesgarani(2019)}]{luo2019conv}
Luo, Y.; and Mesgarani, N. 2019.
\newblock Conv-tasnet: Surpassing ideal time--frequency magnitude masking for
  speech separation.
\newblock \emph{IEEE/ACM transactions on audio, speech, and language
  processing}, 27(8): 1256--1266.

\bibitem[{Manilow et~al.(2019)Manilow, Wichern, Seetharaman, and
  Le~Roux}]{manilow2019}
Manilow, E.; Wichern, G.; Seetharaman, P.; and Le~Roux, J. 2019.
\newblock Cutting Music Source Separation Some {Slakh}: A Dataset to Study the
  Impact of Training Data Quality and Quantity.
\newblock In \emph{Proc. IEEE Workshop on Applications of Signal Processing to
  Audio and Acoustics (WASPAA)}. IEEE.

\bibitem[{Narayanaswamy et~al.(2020)Narayanaswamy, Thiagarajan, Anirudh, and
  Spanias}]{narayanaswamy2020unsupervised}
Narayanaswamy, V.; Thiagarajan, J.~J.; Anirudh, R.; and Spanias, A. 2020.
\newblock Unsupervised Audio Source Separation using Generative Priors.
\newblock arXiv:2005.13769.

\bibitem[{Nugraha, Liutkus, and Vincent(2016)}]{7492604}
Nugraha, A.~A.; Liutkus, A.; and Vincent, E. 2016.
\newblock Multichannel Audio Source Separation With Deep Neural Networks.
\newblock \emph{IEEE/ACM Transactions on Audio, Speech, and Language
  Processing}, 24(9): 1652--1664.

\bibitem[{Parisi(1981)}]{parisi}
Parisi, G. 1981.
\newblock Correlation functions and computer simulations.
\newblock \emph{Nuclear Physics B}, 180(3): 378--384.

\bibitem[{Postolache et~al.(2022)Postolache, Pons, Pascual, and
  Serr{\`a}}]{postolache2022adversarial}
Postolache, E.; Pons, J.; Pascual, S.; and Serr{\`a}, J. 2022.
\newblock Adversarial Permutation Invariant Training for Universal Sound
  Separation.
\newblock \emph{arXiv preprint arXiv:2210.12108}.

\bibitem[{Rafii and Pardo(2012)}]{repet}
Rafii, Z.; and Pardo, B. 2012.
\newblock Repeating pattern extraction technique (REPET): A simple method for
  music/voice separation.
\newblock \emph{IEEE transactions on audio, speech, and language processing},
  21(1): 73--84.

\bibitem[{Ramesh et~al.(2021)Ramesh, Pavlov, Goh, Gray, Voss, Radford, Chen,
  and Sutskever}]{ramesh2021zero}
Ramesh, A.; Pavlov, M.; Goh, G.; Gray, S.; Voss, C.; Radford, A.; Chen, M.; and
  Sutskever, I. 2021.
\newblock Zero-shot text-to-image generation.
\newblock In \emph{Proc. ICML}, 8821--8831. PMLR.

\bibitem[{Razavi, van~den Oord, and Vinyals(2019)}]{razavi2019}
Razavi, A.; van~den Oord, A.; and Vinyals, O. 2019.
\newblock Generating Diverse High-Fidelity Images with {VQ-VAE-2}.
\newblock In \emph{Proc. NeurIPS}.

\bibitem[{Reddy et~al.(1977)}]{reddy1977speechBeam}
Reddy, D.~R.; et~al. 1977.
\newblock Speech understanding systems: A summary of results of the five-year
  research effort.
\newblock \emph{Department of Computer Science. Camegie-Mell University,
  Pittsburgh, PA}, 17: 138.

\bibitem[{Roweis(2000)}]{Roweis:2000}
Roweis, S.~T. 2000.
\newblock One Microphone Source Separation.
\newblock In \emph{Proc. NIPS}.

\bibitem[{Salimans et~al.(2017)Salimans, Karpathy, Chen, and
  Kingma}]{salimans2017pixelcnn++}
Salimans, T.; Karpathy, A.; Chen, X.; and Kingma, D.~P. 2017.
\newblock Pixelcnn++: Improving the pixelcnn with discretized logistic mixture
  likelihood and other modifications.
\newblock \emph{arXiv preprint arXiv:1701.05517}.

\bibitem[{Sanh et~al.(2022)Sanh, Webson, Raffel et~al.}]{sanh2022multitask}
Sanh, V.; Webson, A.; Raffel, C.; et~al. 2022.
\newblock Multitask Prompted Training Enables Zero-Shot Task Generalization.
\newblock In \emph{Proc. ICLR}.

\bibitem[{Seetharaman, Pishdadian, and Pardo(2017)}]{ft2d}
Seetharaman, P.; Pishdadian, F.; and Pardo, B. 2017.
\newblock Music/voice separation using the 2d fourier transform.
\newblock In \emph{2017 IEEE Workshop on Applications of Signal Processing to
  Audio and Acoustics (WASPAA)}, 36--40. IEEE.

\bibitem[{{Smaragdis} et~al.(2014){Smaragdis}, {Févotte}, {Mysore},
  {Mohammadiha}, and {Hoffman}}]{Smaragdis:2014}
{Smaragdis}, P.; {Févotte}, C.; {Mysore}, G.~J.; {Mohammadiha}, N.; and
  {Hoffman}, M. 2014.
\newblock Static and Dynamic Source Separation Using Nonnegative
  Factorizations: A unified view.
\newblock \emph{IEEE Signal Processing Magazine}, 31(3): 66--75.

\bibitem[{Song and Ermon(2019)}]{song2019generative}
Song, Y.; and Ermon, S. 2019.
\newblock Generative Modeling by Estimating Gradients of the Data Distribution.
\newblock In \emph{Advances in Neural Information Processing Systems},
  11895--11907.

\bibitem[{Song et~al.(2021)Song, Sohl-Dickstein, Kingma, Kumar, Ermon, and
  Poole}]{song2021scorebased}
Song, Y.; Sohl-Dickstein, J.; Kingma, D.~P.; Kumar, A.; Ermon, S.; and Poole,
  B. 2021.
\newblock Score-Based Generative Modeling through Stochastic Differential
  Equations.
\newblock In \emph{Proc. ICLR}.

\bibitem[{St{\"o}ter, Liutkus, and Ito(2018)}]{SiSEC18}
St{\"o}ter, F.-R.; Liutkus, A.; and Ito, N. 2018.
\newblock The 2018 Signal Separation Evaluation Campaign.
\newblock In \emph{Proc. LVA/ICA}, 293--305.

\bibitem[{Subakan and Smaragdis(2018)}]{subakan2018generative}
Subakan, Y.~C.; and Smaragdis, P. 2018.
\newblock Generative adversarial source separation.
\newblock In \emph{Proc. ICASSP}, 26--30. IEEE.

\bibitem[{Takahashi, Goswami, and Mitsufuji(2018)}]{Takashi2018}
Takahashi, N.; Goswami, N.; and Mitsufuji, Y. 2018.
\newblock Mmdenselstm: An Efficient Combination of Convolutional and Recurrent
  Neural Networks for Audio Source Separation.
\newblock In \emph{Proc. IWAENC}, 106--110.

\bibitem[{Takahashi et~al.(2019)Takahashi, Parthasaarathy, Goswami, and
  Mitsufuji}]{takahashi2019recursive}
Takahashi, N.; Parthasaarathy, S.; Goswami, N.; and Mitsufuji, Y. 2019.
\newblock Recursive speech separation for unknown number of speakers.
\newblock \emph{arXiv preprint arXiv:1904.03065}.

\bibitem[{Uhlich, Giron, and Mitsufuji(2015)}]{Uhlich:2015}
Uhlich, S.; Giron, F.; and Mitsufuji, Y. 2015.
\newblock Deep neural network based instrument extraction from music.
\newblock In \emph{Proc. ICASSP}.

\bibitem[{van~den Oord et~al.(2016)van~den Oord, Kalchbrenner, Espeholt,
  Vinyals, Graves et~al.}]{van2016conditional}
van~den Oord, A.; Kalchbrenner, N.; Espeholt, L.; Vinyals, O.; Graves, A.;
  et~al. 2016.
\newblock Conditional image generation with pixelcnn decoders.
\newblock \emph{Proc. NeurIPS}, 29.

\bibitem[{van~den Oord, Vinyals, and Kavukcuoglu(2017)}]{oord2017}
van~den Oord, A.; Vinyals, O.; and Kavukcuoglu, K. 2017.
\newblock Neural Discrete Representation Learning.
\newblock In \emph{Proc. NeurIPS}.

\bibitem[{Vaswani et~al.(2017)Vaswani, Shazeer, Parmar, Uszkoreit, Jones,
  Gomez, Kaiser, and Polosukhin}]{vaswani2017attention}
Vaswani, A.; Shazeer, N.; Parmar, N.; Uszkoreit, J.; Jones, L.; Gomez, A.~N.;
  Kaiser, {\L}.; and Polosukhin, I. 2017.
\newblock Attention is all you need.
\newblock \emph{Proc. NeurIPS}, 30.

\bibitem[{Wei et~al.(2021)Wei, Bosma, Zhao, Guu, Yu, Lester, Du, Dai, and
  Le}]{weiFLAN}
Wei, J.; Bosma, M.; Zhao, V.~Y.; Guu, K.; Yu, A.~W.; Lester, B.; Du, N.; Dai,
  A.~M.; and Le, Q.~V. 2021.
\newblock Finetuned Language Models Are Zero-Shot Learners.
\newblock \emph{CoRR}, abs/2109.01652.

\bibitem[{Wisdom et~al.(2021)Wisdom, Erdogan, Ellis, Serizel, Turpault,
  Fonseca, Salamon, Seetharaman, and Hershey}]{fuss}
Wisdom, S.; Erdogan, H.; Ellis, D. P.~W.; Serizel, R.; Turpault, N.; Fonseca,
  E.; Salamon, J.; Seetharaman, P.; and Hershey, J.~R. 2021.
\newblock What's all the Fuss about Free Universal Sound Separation Data?
\newblock In \emph{Proc. ICASSP}, 186--190.

\bibitem[{Wisdom et~al.(2020)Wisdom, Tzinis, Erdogan, Weiss, Wilson, and
  Hershey}]{mixit}
Wisdom, S.; Tzinis, E.; Erdogan, H.; Weiss, R.~J.; Wilson, K.~W.; and Hershey,
  J.~R. 2020.
\newblock Unsupervised Sound Separation Using Mixture Invariant Training.

\bibitem[{Xu et~al.(2021)Xu, Song, Garg, Gong, Shu, Grover, and
  Ermon}]{xu2021anytime}
Xu, Y.; Song, Y.; Garg, S.; Gong, L.; Shu, R.; Grover, A.; and Ermon, S. 2021.
\newblock Anytime sampling for autoregressive models via ordered autoencoding.
\newblock \emph{arXiv preprint arXiv:2102.11495}.

\bibitem[{Yang et~al.(2022)Yang, Lin, Yang, Wang, Zhou, and
  Yang}]{https://doi.org/10.48550/arxiv.2208.02532}
Yang, H.; Lin, J.; Yang, A.; Wang, P.; Zhou, C.; and Yang, H. 2022.
\newblock Prompt Tuning for Generative Multimodal Pretrained Models.

\bibitem[{Yu et~al.(2021)Yu, Li, Koh, Zhang, Pang, Qin, Ku, Xu, Baldridge, and
  Wu}]{yu2021vector}
Yu, J.; Li, X.; Koh, J.~Y.; Zhang, H.; Pang, R.; Qin, J.; Ku, A.; Xu, Y.;
  Baldridge, J.; and Wu, Y. 2021.
\newblock Vector-quantized image modeling with improved vqgan.
\newblock \emph{arXiv preprint arXiv:2110.04627}.

\bibitem[{Yu et~al.(2022)Yu, Xu, Koh, Luong, Baid, Wang, Vasudevan, Ku, Yang,
  Ayan et~al.}]{yu2022scaling}
Yu, J.; Xu, Y.; Koh, J.~Y.; Luong, T.; Baid, G.; Wang, Z.; Vasudevan, V.; Ku,
  A.; Yang, Y.; Ayan, B.~K.; et~al. 2022.
\newblock Scaling autoregressive models for content-rich text-to-image
  generation.
\newblock \emph{arXiv preprint arXiv:2206.10789}.

\end{thebibliography}

\end{document}